\documentclass{article}

\usepackage[preprint]{neurips_2025}
\makeatletter
\renewcommand{\@notice}{}
\makeatother

\usepackage[utf8]{inputenc}
\usepackage[T1]{fontenc}
\usepackage{hyperref}
\usepackage{url}
\usepackage{booktabs}
\usepackage{amsfonts}
\usepackage{nicefrac}
\usepackage{microtype}
\usepackage{xcolor}
\usepackage{amsmath}
\usepackage{amssymb}
\usepackage{multirow}
\usepackage{subcaption}
\usepackage{graphicx}
\usepackage{tabularx}
\usepackage{array}
\usepackage{siunitx}
\usepackage{lmodern}
\usepackage{enumitem}
\usepackage{tikz}
\usetikzlibrary{arrows.meta, positioning, calc, fit, backgrounds, shadows.blur}
\tikzset{
  figbox/.style={draw=blue!60, fill=blue!10, rounded corners=3pt, align=center, font=\small},
  figghost/.style={draw=gray!40, fill=gray!6, rounded corners=3pt, align=center, font=\small, text=gray!70},
  figmtp/.style={draw=orange!75!black, fill=orange!12, rounded corners=3pt, align=center, font=\small},
  figarr/.style={-{Latex[length=2mm]}, thick, black!55, shorten >=0.6pt, shorten <=0.6pt},
  figflow/.style={-{Latex[length=2mm]}, thick, blue!65, shorten >=0.6pt, shorten <=0.6pt},
  figoarr/.style={-{Latex[length=2mm]}, thick, orange!75!black, shorten >=0.6pt, shorten <=0.6pt},
  figlbl/.style={font=\scriptsize, text=black!70, inner sep=1.5pt, align=center},
  figtag/.style={font=\scriptsize, text=blue!65, inner sep=1.5pt},
}
\usepackage{float}

\usepackage{adjustbox}

\makeatletter
\def\@fnsymbol#1{\ensuremath{\ifcase#1\or \dagger\or \ddagger\or \S\or \|%
  \or **\or \dagger\dagger\or \ddagger\ddagger \else\@ctrerr\fi}}
\makeatother

\newcommand{\jinaocr}{\href{https://huggingface.co/jinaai/jina-ocr-v1}{\texttt{Jina-OCR-v1}}}

\title{\jinaocr{}: Efficient Document Parsing with\\ Speculative Decoding and Dense Verifiable Rewards}

\author{%
  \textbf{Alejandro Barón García} \quad \textbf{Feng Wang} \\
  \textbf{Emilia Garcia Casademont} \quad \textbf{Han Xiao} \\
  \\
  Jina AI \textit{by} Elastic \\
  33 New Montgomery Street, San Francisco, CA 94105, USA \\
  \texttt{research@jina.ai} \\
}

\begin{document}
\maketitle

\begin{abstract}
We present \jinaocr{}, an end-to-end document parsing model built to
serve on low-budget GPUs. It combines the compressed-vision encoder and the
3B mixture-of-experts decoder of DeepSeek-OCR, which activates about 570M
parameters per token, with a FastMTP speculative decoding head that shares a
single draft block recursively across $K{=}3$ prediction steps. Greedy
verification makes decoding lossless. Post-training combines instruction
alignment, robustness fine-tuning on difficult documents, and GRPO under
dense verifiable rewards: deterministic formula, table, and structural
checks that award partial credit. The training data mixes cleaned public
corpora with targeted synthetic pages. At the default
dynamic-resolution setting, \jinaocr{} scores 91.14 on OmniDocBench v1.6 and 83.4 on
olmOCR-Bench, and reaches the highest page throughput in our comparison at
2.57 pages per second. On a low-budget GPU such as the NVIDIA L4, FastMTP
doubles decoding speed over greedy autoregressive decoding. The model
is publicly available at \url{https://huggingface.co/jinaai/jina-ocr-v1}.
\end{abstract}

\section{Introduction}
\label{sec:introduction}

End-to-end document parsing with vision-language models (VLMs) has recently
emerged as a prominent paradigm
\citep{wei2024got,nassar2025smoldocling,rednote2025dots,wei2025deepseekocr},
collapsing layout analysis, text recognition, formula transcription, and table
structure recovery into a single generative pass. Its outputs provide
machine-readable representations for retrieval, grounding, and agentic
workflows, but production use, especially on the low-budget GPUs common
in deployment, remains constrained by decoding cost and post-training data
coverage.

Long outputs make decoding expensive. General-purpose VLMs such as
Qwen2.5-VL-72B \citep{bai2025qwen25vl} use thousands of vision tokens per
page and decode strictly autoregressively. DeepSeek-OCR
\citep{wei2025deepseekocr} showed that a compressed vision encoder and a
compact MoE decoder substantially reduce the serving cost. We build on that
architecture and target the remaining autoregressive bottleneck: the
local structure of OCR output makes it possible to draft several tokens
before the verifier evaluates them.

Post-training requires both reliable supervision and coverage. Open labels
may contain degeneration loops or malformed structures, and formula- and
table-specific rewards apply only to pages containing those elements. At
the same time, document requests span full-page parsing, element
transcription, and different formatting conventions.

Our principal design choices are:

\begin{itemize}[leftmargin=1.5em,itemsep=2pt,topsep=2pt]
  \item \textbf{FastMTP decoding.} We attach a recursively shared FastMTP
  \citep{fastmtp2025} draft block ($K{=}3$) to the compressed-vision, 3B-MoE
  backbone of DeepSeek-OCR. Sharing one block across prediction depths keeps
  the draft parameter count constant in $K$.
  \item \textbf{Post-training under dense verifiable rewards.} We optimize a
  multiplicative GRPO reward over content, structure, unit tests,
  repetition, and format with a ReMax baseline \citep{li2023remax}. Every
  term is deterministic code against a reference and is graded, so a
  partially correct page receives partial credit. Synthetic pages packed with scorable
  formulas and tables (JinaOCRSynth) raise the share of training pages on
  which the formula and table terms apply.
  \item \textbf{Instruction-oriented training.} Building on ReaderLM-v2
  \citep{wang2025readerlmv2}, the mixture covers multiple parsing styles,
  element-level transcription, captioning, document VQA, and
  key-information extraction, with dynamically generated instructions and
  auxiliary grounding examples.
\end{itemize}

\begin{figure}[t]
  \centering
  \includegraphics[width=\textwidth]{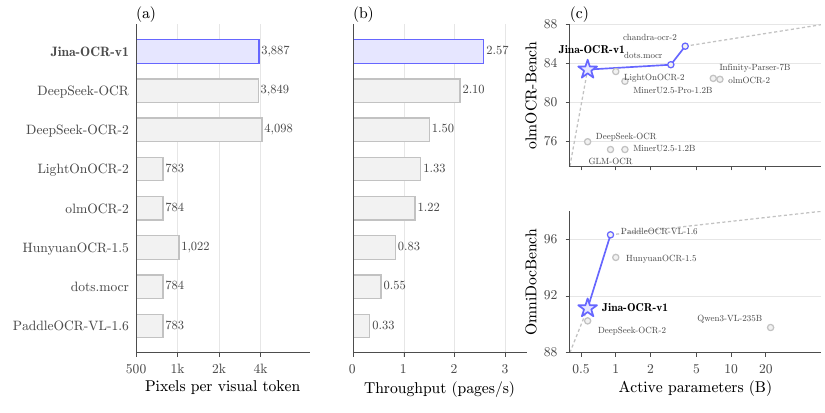}
  \caption{Overview of specialized OCR models; blue marks this work.
    (a)~Pixels per visual token (log scale) with the measured value.
    (b)~Page throughput on olmOCR-Bench (pages/s, one A100, concurrency 32).
    (c)~Benchmark overall against active parameters (log scale) on
    olmOCR-Bench (top) and OmniDocBench (bottom); the solid line joins the
    Pareto-optimal systems and the dashed segments extend it to the axis
    corners; unlabeled points are the remaining systems of
    \autoref{tab:olmocr}, \autoref{tab:main}, and \autoref{tab:efficiency}.}
  \label{fig:overview}
\end{figure}

\autoref{fig:overview} places \jinaocr{} on three axes that jointly
determine deployment cost: visual compression, page throughput,
and accuracy per active parameter.

Panel~(a) measures how aggressively the vision encoder compresses a page.
The effective patch side $p$ is the spatial extent of one visual token, and
\emph{pixels per visual token} is how many image pixels that token
represents. A square-patch ViT with patch side $p$ yields about $p^{2}$
pixels per token, so the $28$--$32$~px encoders used by the Qwen-VL
family and related models sit near $780$--$1{,}000$ pixels/token.
DeepEncoder instead maps a $1024{\times}1024$ global view from $4{,}096$
patches to $256$ tokens ($p{=}64$, about $4{,}096$ pixels/token).
\jinaocr{} inherits that encoder, so a page of comparable resolution is
represented by far fewer visual tokens.

Panel~(b) reports page throughput on olmOCR-Bench. Visual compression alone
does not determine it: DeepSeek-OCR matches the pixel-per-token density of
\jinaocr{} while emitting longer pages, and \autoref{sec:eval-efficiency}
measures the resulting difference.

Panel~(c) shows that parsing quality is preserved.
\jinaocr{} activates $570$M decoder parameters per token and lies on the
accuracy frontier against that budget on both benchmarks, ahead of systems
an order of magnitude larger, and on OmniDocBench it is the most compact
point on that frontier.

\autoref{sec:related} reviews related work. \autoref{sec:architecture}
specifies the FastMTP head, \autoref{sec:data} the training data,
\autoref{sec:training} the post-training recipe, and
\autoref{sec:evaluation} the evaluation.

\section{Related Work}
\label{sec:related}

\paragraph{End-to-end document parsing.}
Nougat \citep{blecher2023nougat} first demonstrated end-to-end academic
document OCR, and GOT-OCR2.0 \citep{wei2024got} extended the scope to a broad
OCR-2.0 task suite. Recent systems include dots.ocr \citep{rednote2025dots},
SmolDocling \citep{nassar2025smoldocling}, olmOCR \citep{poznanski2025olmocr},
MonkeyOCR and MonkeyOCRv2 \citep{li2025monkeyocr,liu2026monkeyocrv2},
MinerU \citep{wang2024mineru}, PaddleOCR-VL \citep{cui2025paddleocr},
GLM-OCR \citep{duan2026glmocr}, Infinity-Parser
\citep{infteam2026infinityparser2}, and DeepSeek-OCR
\citep{wei2025deepseekocr}, which is the architectural base of this work.
The list includes both end-to-end systems and two-stage pipelines that first
run a layout model such as PP-DocLayout
\citep{sun2025ppdoclayoutunifieddocumentlayout}.

\paragraph{Efficient decoding and multi-token prediction.}
Speculative decoding \citep{leviathan2023speculative} verifies drafted tokens
in parallel. The idea of predicting several tokens from parallel heads goes
back to blockwise parallel decoding \citep{stern2018blockwise}; Medusa
\citep{cai2024medusa} revived it as a set of independent draft heads on a
frozen backbone, and Hydra \citep{ankner2024hydra} made those heads
sequentially dependent. Lookahead decoding \citep{fu2024lookahead} removes
the draft model entirely. Multi-token prediction was popularized as a
training objective by \citet{gloeckle2024multitoken} and DeepSeek-V3
\citep{liu2024deepseekv3}; EAGLE-style methods
\citep{li2024eagle,li2025eagle3} draft at the feature level, and
\citet{draftkl2026} train draft heads on the acceptance rate directly.
FastMTP \citep{fastmtp2025} shares one draft block across prediction
depths. GLM-OCR \citep{duan2026glmocr} already applies shared-parameter MTP
to OCR, and HunyuanOCR-1.5 \citep{hunyuanocr15} adopts DFlash block drafting
\citep{chen2026dflash}. We follow the shared-block line for its constant
draft parameter count and train it with feature- and distribution-alignment
losses.

\paragraph{Reinforcement learning for document parsing.}
GRPO \citep{shao2024deepseekmath} and DAPO \citep{yu2025dapo} have been
adapted to document parsing under reinforcement learning with verifiable
rewards \citep[RLVR;][]{lambert2025tulu3}, using signals such as edit
distance, TEDS \citep{zhong2020teds}, and CDM \citep{cdm2024}. The
structured content of document elements makes them natural targets for RLVR,
since a reference transcription admits deterministic scoring without a
reward model or a judge. RL is also less prone to catastrophic forgetting
than supervised fine-tuning \citep{shenfeld2026rls}, which supports
specializing on sub-tasks through different rewards. LightOnOCR and the
olmOCR project use unit-test-style evaluation and GRPO over document
training \citep{lightonocr,poznanski2026olmocrproject}, and
Infinity-Parser2 \citep{infteam2026infinityparser2} co-trains eight tasks
under a multi-task reward. We combine a multiplicative reward with per-term
floors, a ReMax baseline \citep{li2023remax}, an explicit repetition
penalty, and synthetic pages concentrated on formula and table structures.

\section{Model Architecture}
\label{sec:architecture}

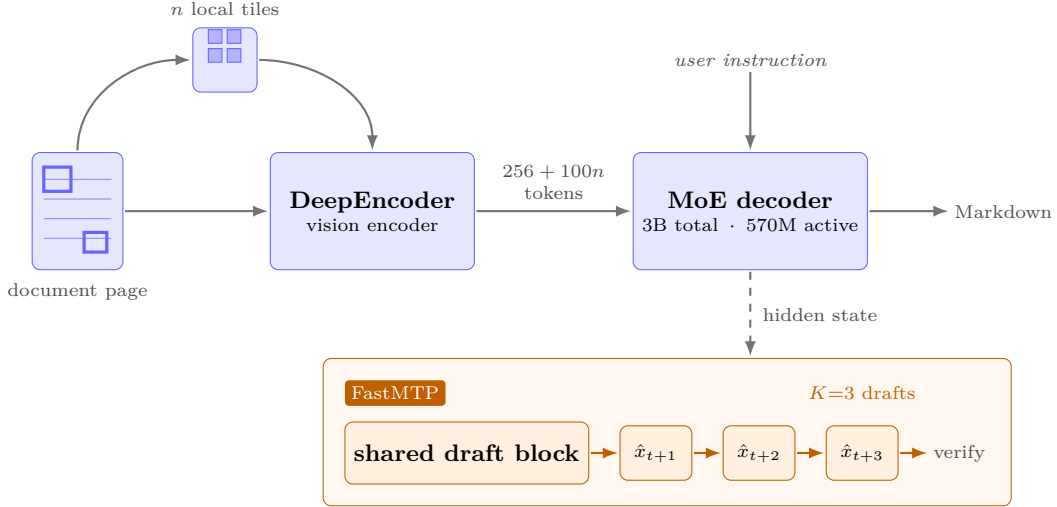
\begin{figure}[t]
  \centering
  \begin{tikzpicture}[every node/.style={align=center}]
    \node[figbox, minimum width=1.2cm, minimum height=1.55cm] (page) at (0,0) {};
    \foreach \yy in {0.42,0.16,-0.10,-0.36}
      \draw[blue!45, thin] ($(page.west)+(0.16,\yy)$) -- ($(page.east)+(-0.16,\yy)$);
    \draw[blue!60, very thick] ($(page.north west)+(0.16,-0.20)$) rectangle ++(0.36,-0.32);
    \draw[blue!60, very thick] ($(page.south east)+(-0.52,0.24)$) rectangle ++(0.30,0.26);
    \node[figlbl, below=3pt] at (page.south) {document page};

    \node[figbox, minimum width=0.85cm, minimum height=0.85cm] (tiles) at (1.95,2.0) {};
    \foreach \xx/\yy in {-0.22/0.22, 0.03/0.22, -0.22/-0.03, 0.03/-0.03}
      \draw[blue!60, thin, fill=blue!30] ($(tiles.center)+(\xx,\yy)$) rectangle ++(0.18,0.18);
    \node[figlbl, above=3pt] at (tiles.north) {$n$ local tiles};

    \node[figbox, minimum width=2.7cm, minimum height=1.55cm] (enc) at (3.9,0)
      {\textbf{DeepEncoder}\\[-1pt] {\scriptsize vision encoder}};
    \node[figbox, minimum width=3.05cm, minimum height=1.55cm] (dec) at (8.9,0)
      {\textbf{MoE decoder}\\[-1pt] {\scriptsize 3B total \,$\cdot$\, 570M active}};
    \node[figlbl] (out) at (12.25,0) {Markdown};
    \node[figlbl, font=\scriptsize\itshape] (instr) at (8.9,2.0) {user instruction};

    \node[figmtp, minimum width=2.35cm, minimum height=0.82cm] (draft) at (5.15,-3.2)
      {\textbf{shared draft block}};
    \node[figmtp, minimum width=0.95cm, minimum height=0.72cm, right=0.4cm of draft] (t1) {\scriptsize $\hat{x}_{t+1}$};
    \node[figmtp, minimum width=0.95cm, minimum height=0.72cm, right=0.4cm of t1] (t2) {\scriptsize $\hat{x}_{t+2}$};
    \node[figmtp, minimum width=0.95cm, minimum height=0.72cm, right=0.4cm of t2] (t3) {\scriptsize $\hat{x}_{t+3}$};
    \node[figlbl, right=0.4cm of t3] (ver) {verify};
    \node[figlbl, text=orange!75!black, anchor=south] (klabel) at ($(t3.north)+(0,8pt)$) {$K{=}3$ drafts};
    \node[font=\scriptsize, text=white, fill=orange!75!black, rounded corners=1.5pt,
          inner sep=2.4pt] (tag) at (draft.west |- klabel) [anchor=west] {FastMTP};

    \begin{scope}[on background layer]
      \node[figmtp, fill=orange!6, fit=(draft)(t1)(t2)(t3)(ver)(klabel)(tag), inner sep=8pt] (panel) {};
    \end{scope}

    \draw[figarr] (page) -- (enc);
    \draw[figarr] (page.north) to[out=90,in=180] (tiles.west);
    \draw[figarr] (tiles.east) to[out=0,in=90] (enc.north);
    \draw[figarr] (enc) -- (dec);
    \node[figlbl, above=3pt] at ($(enc.east)!0.5!(dec.west)$) {$256+100n$\\tokens};
    \draw[figarr] (dec) -- (out);
    \draw[figarr] (instr.south) -- (dec.north);
    \draw[figarr, dashed] (dec.south) -- (dec.south |- panel.north);
    \node[figlbl, right=3pt] at ($(dec.south)!0.5!(dec.south |- panel.north)$) {hidden state};
    \draw[figoarr] (draft) -- (t1);
    \draw[figoarr] (t1) -- (t2);
    \draw[figoarr] (t2) -- (t3);
    \draw[figoarr] (t3) -- (ver);
  \end{tikzpicture}
  \caption{Architecture of \jinaocr{}. DeepEncoder and the MoE decoder follow
  DeepSeek-OCR; a page yields a $1024{\times}1024$ global view of 256 visual
  tokens plus $n$ local tiles of 100 tokens each. The FastMTP head
  \citep{fastmtp2025} (orange) recursively proposes $K{=}3$ tokens from one
  shared draft block for verifier evaluation.}
  \label{fig:arch}
\end{figure}

\jinaocr{} follows the encoder--decoder architecture of DeepSeek-OCR
\citep{wei2025deepseekocr} and extends it with a multi-token prediction head.
\autoref{fig:arch} illustrates the overall system, and
\autoref{tab:arch} summarizes both the inherited backbone and the added
FastMTP component.

\begin{table}[t]
  \centering
  \caption{Model specification of \jinaocr{}. DeepEncoder and the MoE
  decoder are inherited from DeepSeek-OCR; FastMTP is added on top.}
  \label{tab:arch}
  \footnotesize
  \begin{tabular}{ll}
    \toprule
    Component & Specification \\
    \midrule
    Vision encoder & DeepEncoder ($\sim$380M): SAM (80M) $\to$ 16$\times$ conv $\to$ CLIP-L (300M) \\
    Vision tokens & 256 @ $1024{\times}1024$ (Base); $256{+}100n$, $n{\leq}9$ (Gundam, $\leq$1{,}156/page) \\
    Decoder & DeepSeek-3B-MoE: 12 layers, $d{=}1280$, 64 routed + 2 shared, top-6 \\
    Active / total params & $\sim$570M / $\sim$3B (decoder); $<$1B / $\sim$3.4B (whole model) \\
    Vocabulary & 129{,}280 \\
    Position limit & 32{,}768 (RoPE \citep{su2021rope}, $\theta{=}10^6$) \\
    MTP head & 1 shared dense block, recursive $K{=}3$ steps (FastMTP) \\
    \bottomrule
  \end{tabular}
\end{table}

DeepEncoder cascades a window-attention SAM encoder \citep{kirillov2023sam}
and a global-attention CLIP encoder \citep{radford2021clip} through a
16$\times$ convolutional compressor. At $1024{\times}1024$ it compresses
4{,}096 global patches to 256 tokens, and the dynamic-resolution Gundam
mode adds at most nine 100-token local tiles. The decoder is a compact DeepSeekMoE model
\citep{dai2024deepseekmoe,liu2024deepseekv2} that activates approximately
570M parameters per token and emits Markdown. These components provide the
compressed visual prefix and the verifier used by FastMTP; the remainder of
this section focuses on the added draft head.

\subsection{Draft-head architecture}
\label{sec:fastmtp}

OCR output is near-deterministic with strong local structure, which makes it
a favorable speculative-decoding workload. We adopt FastMTP
\citep{fastmtp2025}: a \emph{single} draft block $B_\theta$ is applied
recursively for $K{=}3$ steps, so the draft parameter count is constant in
$K$.

Training aligns the recursive depths on a two-dimensional grid of sequence
positions and depths. Let $x_p$ be
the token at sequence position $p$, $e_p=E(x_p)$ its embedding, and $b_p$
the post-norm state of the main decoder after consuming $x_p$. Let
$u_p^{(k)}$ denote the draft output at training row $p$ and recursive depth
$k$. The state presented to the shared head is
\begin{equation}
\begin{aligned}
  a_p^{(1)} &= b_p, &
  a_p^{(k)} &= u_{p-1}^{(k-1)} \quad (k>1),\\
  z_p^{(k)} &= P\!\left([\,N_e(e_{p+1});N_h(a_p^{(k)})\,]\right),\\
  u_p^{(k)} &= N_o\!\left(B_\theta(z_p^{(k)};\operatorname{pos}=p)\right),&
  \ell_p^{(k)} &= W_{\mathrm{lm}}u_p^{(k)} .
\end{aligned}
\end{equation}
Here $N_e$ and $N_h$ are normalization layers, $P$ is a linear projection,
and $B_\theta$ is the shared dense draft block.\footnote{In the released
checkpoint these are \texttt{enorm}, \texttt{hnorm}, \texttt{eh\_proj},
and \texttt{mtp\_block}.} The token embedding $E$,
output norm $N_o$, and LM head $W_{\mathrm{lm}}$ are tied to their fixed
verifier counterparts. Embeddings and RoPE positions remain fixed across
depths, whereas the predicted state shifts from row $p-1$ to row $p$.
Prompt rows keep their base states at every depth, matching the
committed-token KV cache used at inference.

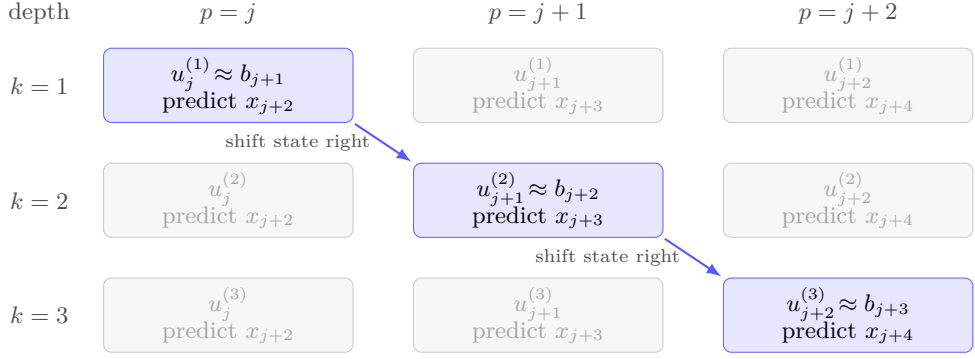
\begin{figure}[t]
  \centering
  \begin{tikzpicture}[
    cell/.style={minimum width=3.3cm, minimum height=0.72cm},
    active/.style={figbox, cell},
    ghost/.style={figghost, cell},
  ]
    \node[figlbl, font=\small] at (-2.5,0) {depth};
    \node[figlbl, font=\small] at (0,0) {$p=j$};
    \node[figlbl, font=\small] at (4.1,0) {$p=j+1$};
    \node[figlbl, font=\small] at (8.2,0) {$p=j+2$};

    \node[figlbl, font=\small] at (-2.5,-0.95) {$k=1$};
    \node[active] (s1) at (0,-0.95) {$u_j^{(1)}\!\approx b_{j+1}$\\predict $x_{j+2}$};
    \node[ghost] at (4.1,-0.95) {$u_{j+1}^{(1)}$\\predict $x_{j+3}$};
    \node[ghost] at (8.2,-0.95) {$u_{j+2}^{(1)}$\\predict $x_{j+4}$};

    \node[figlbl, font=\small] at (-2.5,-2.47) {$k=2$};
    \node[ghost] at (0,-2.47) {$u_j^{(2)}$\\predict $x_{j+2}$};
    \node[active] (s2) at (4.1,-2.47) {$u_{j+1}^{(2)}\!\approx b_{j+2}$\\predict $x_{j+3}$};
    \node[ghost] at (8.2,-2.47) {$u_{j+2}^{(2)}$\\predict $x_{j+4}$};

    \node[figlbl, font=\small] at (-2.5,-3.99) {$k=3$};
    \node[ghost] at (0,-3.99) {$u_j^{(3)}$\\predict $x_{j+2}$};
    \node[ghost] at (4.1,-3.99) {$u_{j+1}^{(3)}$\\predict $x_{j+3}$};
    \node[active] (s3) at (8.2,-3.99) {$u_{j+2}^{(3)}\!\approx b_{j+3}$\\predict $x_{j+4}$};

    \draw[figflow] (s1.south east) -- node[figlbl, left=3pt] {shift state right} (s2.north west);
    \draw[figflow] (s2.south east) -- node[figlbl, left=3pt] {shift state right} (s3.north west);
  \end{tikzpicture}
  \caption{FastMTP state-shift alignment. Training computes every row in
  parallel; within a fixed column, all depths use the same token and verifier
  targets. The highlighted diagonal is one inference-time draft chain, whose
  target advances by one token at each recursive application.}
  \label{fig:mtp-shift}
\end{figure}

For every depth, training row $p$ estimates the same verifier state
$b_{p+1}$ and its logits predict $x_{p+2}$. The apparent multi-token advance
is diagonal: for a draft chain beginning at row $j$, depth $k$ occupies
$p=j+k-1$, so it predicts $x_{j+k+1}$. \autoref{fig:mtp-shift} makes this
train--inference correspondence explicit.

At inference we follow the EAGLE \citep{li2024eagle} feature-level drafting
paradigm, in which the draft head proposes tokens from hidden states as well
as token ids. The speculative-decoding loop
\citep{leviathan2023speculative} uses greedy verification. Let
$\hat{x}_{1:K}$ be the draft candidates and
$v_r=\arg\max_x q_r(x)$ the verifier token at continuation position $r$.
We accept the longest equality prefix
\begin{equation}
  m=\max\bigl(\{0\}\cup
  \{r\leq K:\hat{x}_s=v_s\ \text{for all}\ s\leq r\}\bigr).
\end{equation}
The decoder commits $\hat{x}_{1:m}$ followed by $v_{m+1}$. On a mismatch,
the greedy token of the verifier replaces the first rejected draft and the
remaining candidates are discarded; when all $K$ candidates match,
$v_{K+1}$ is committed as a bonus token. No stochastic acceptance or
residual-distribution resampling is used. Decoding is therefore lossless:
for any acceptance pattern, the committed sequence is the greedy sequence
the verifier would have produced on its own. \autoref{sec:stage4}
specifies the training objective of the head.

\section{Training Data}
\label{sec:data}

The training data addresses label quality and the availability of structures
covered by the reward functions. We assemble and clean a mixture of public
corpora, synthesize pages where open data is sparse or difficult to
score, and pair every sample with a task instruction, with collection
guided by model error analysis. \autoref{tab:data} gives an overview.

\subsection{Open data collection and cleaning}
\label{sec:data-collect}

The mixture draws from public document OCR corpora including olmOCR-mix and
olmOCR-synthmix \citep{poznanski2025olmocr}, FinePDFs \citep{finepdfs2025},
LightOnOCR \citep{lightonocr}, DoclingMatrix, RVL-CDIP
\citep{harley2015rvlcdip}, DocLocal4K \citep{hu2024docowl}, SynthChartNet
\citep{nassar2025smoldocling}, MMTab \citep{zheng2024mmtab}, \LaTeX{}-OCR,
UniMER, CommonForms, and the VDR multi-domain corpus \citep{vdr2025}. For
difficult real-world pages, we add historical and degraded sources:
Europeana newspapers \citep{livingmachines_lnnewspapers}, Library of
Congress transcripts \citep{revolutioncrossroads_loc}, NARA pension files
\citep{revolutioncrossroads_nara}, NCSE periodicals \citep{ncse_ucl}, and
Japanese wild-text samples \citep{jawildtext}. All corpora are converted
into a unified instruction--response format with page images.

Cleaning combines two filters. The first is rule-based: a detector
inspects the last 1{,}024 characters of long labels for exact,
truncation-tolerant, and gapped repeats and drops the degeneration loops it
finds; exact duplicate images and samples shorter than 64 tokens are also
discarded.

The second relabels and audits selected corpora with a vision-language
model. Qwen3.5-122B-A10B \citep{qwen2025} relabels hard sources such as
historical newspapers, archives, and forms, and Qwen3-VL-30B-A3B audits
generated or noisy samples.

Collection is guided by error analysis through classifier-driven retrieval.
Following the retrieval strategy of
\citet{5team2025glm45agenticreasoningcoding}, we train modality-specific
classifiers to flag pages associated with a failure mode, using handcrafted
features such as table-structure indicators and distinctive \LaTeX{}
glyphs, model embeddings, or both. Retrieval is approximate and does not
require scoring the current model on every candidate.

\subsection{Targeted synthesis}
\label{sec:data-synth}

Synthesis serves three purposes. The first is layout diversity, for which we
render self-contained HTML documents such as reports, slides, blogs, and
infographics from Wikidata and other openly licensed sources. The second is
difficulty: preference pairs from degraded scans and historical newspapers,
in which a teacher model refines a draft transcription and a judge model
verifies the improvement against the page image; only pairs with a clear
winner are kept.

The third purpose is reward coverage. On naturally distributed pages the
formula- and table-specific reward terms are vacuous for most samples, so a
large share of GRPO rollouts carries no structural signal. JinaOCRSynth
therefore packs each page with scorable structure: a formula suite covering
accent and font confusion, matrix environments, macro conventions, and
unicode glyphs, and a table suite derived from observed failures such as
spanned headers, dense numeric cells, and transposed or borderless layouts.
Each synthetic page carries olmOCR-Bench-style unit tests, which makes the
corresponding reward terms applicable to it.

\subsection{Instruction coverage}
\label{sec:data-instruction}

Training prompts come from a set of task-specific instructions:
full-page parsing in several formatting conventions, element-level transcription of
tables and formulas, image captioning, document VQA, and key-information
extraction, with English and Chinese prompt variants. Each dataset is
paired with one assigned prompt. A subset of samples uses dynamically
generated per-image instructions. Grounding and VQA serve as auxiliary
visual coverage; the deployment target is OCR.

\begin{table}[t]
  \centering
  \caption{Training data composition of \jinaocr{}.}
  \label{tab:data}
  \footnotesize
  \resizebox{\textwidth}{!}{%
  \begin{tabular}{lll}
    \toprule
    Family & Representative sources & Role \\
    \midrule
    Born-digital PDFs & olmOCR-mix, FinePDFs, LightOnOCR, DoclingMatrix & Core parsing \\
    Synthetic pages & olmOCR-synthmix, HTML renderings & Layout diversity \\
    Formula and table & \LaTeX{}-OCR, UniMER, MMTab, SynthChartNet & Structure fidelity \\
    Historical and degraded & Europeana, Library of Congress, NARA, NCSE, jaWildText & Robustness \\
    Multilingual & FinePDFs, Chinese PDFs & Language coverage \\
    Business and forms & Invoices, CommonForms, RVL-CDIP & Key-information extraction \\
    VQA and grounding & DocLocal4K, VDR & Auxiliary visual coverage \\
    Reward-coverage synthesis & JinaOCRSynth (formula and table) & GRPO reward coverage \\
    \bottomrule
  \end{tabular}}
\end{table}

\section{Post-Training}
\label{sec:training}

\autoref{tab:recipe} summarizes the four training objectives used to
develop \jinaocr{}. Alignment SFT, robustness SFT, and GRPO are applied
repeatedly across the rounds of the outer loop in \autoref{sec:merging},
and FastMTP is fit once on the resulting verifier.

\begin{table}[t]
  \centering
  \caption{Training objectives of \jinaocr{}.}
  \label{tab:recipe}
  \footnotesize
  \resizebox{\textwidth}{!}{%
  \begin{tabular}{lllll}
    \toprule
    Objective & Focus & Trainable modules & Learning rate & Sequence length \\
    \midrule
    Alignment SFT & Instructions, length extrapolation & Decoder, projector & $6{\times}10^{-6}$ & 10K, packed \\
    Robustness SFT & Degraded and hard pages & Vision tower, decoder & $10^{-6}$ & 10K \\
    GRPO & Formula and table fidelity & Decoder, projector & per run & 2--4K completion \\
    FastMTP & Draft-head fitting & Draft head & $10^{-4}$--$5{\times}10^{-4}$ & 4--8K \\
    \bottomrule
  \end{tabular}}
\end{table}

\subsection{Alignment SFT and length extrapolation}
\label{sec:stage1}

Alignment SFT makes the base model follow instructions. We train
the language-model decoder, the cross-modal projector, and the output head
with sequence packing and keep the vision tower (SAM, compressor, CLIP)
frozen, since the base visual features are already usable.

Two design choices define this stage. The first extends the usable position
range without long-sequence training. The base model is trained at 8K
context, while dense newspapers and reports need longer outputs, so with
probability 0.4 we shift the position IDs of the \emph{response} tokens by a
random offset drawn from $[1, \min(5L,\, L_{\mathrm{train}}/4)]$, leaving
prompt positions untouched and capping IDs at 32{,}768. Here $L$ is the
tokenized sequence length and $L_{\mathrm{train}}$ is the packed training
length, 10K for alignment SFT. The model therefore sees RoPE
positions beyond the packed length. The second balances the mixture: we
sample across roughly 15 document domains with hierarchical,
domain-balanced rates so that no single source dominates.

\subsection{Robustness fine-tuning}
\label{sec:stage2}

Robustness fine-tuning unfreezes the vision tower at $10^{-6}$ and disables
packing so that long pages stay intact. It combines three mechanisms. On
historical scans we apply photometric and geometric corruptions with
probability 0.25, covering small skews, orthogonal rotations, $2{\times}$
downscaling, salt-and-pepper noise, and the morphological erosion and
dilation that mimic faded print and ink bleed. Labels are then rewritten
toward one formatting convention, so that equations appear individually
delimited as $\$x+y\$,\$t+v\$$ instead of comma-separated as
$\$x+y,t+v\$$, which keeps formatting variation from dominating the loss.
Finally we upweight historical newspaper scans, degraded documents, and the
synthetic formula- and table-heavy pages that carry the most structural
supervision, and oversample the failure modes that error analysis
identifies after each round of the outer loop.

\subsection{GRPO under dense verifiable rewards}
\label{sec:stage3}

Structure-focused GRPO runs use the DAPO variant
\citep{shao2024deepseekmath,yu2025dapo} with eight rollouts per prompt.
Every reward term is computed by deterministic code against a reference
transcription, placing the recipe in the RLVR setting. Temperature, KL
coefficient, completion limit, and baseline policy vary by run. Two
adaptations are specific to OCR.

The first replaces the group-normalized baseline with a ReMax baseline
\citep{li2023remax}, the sampled reward minus the greedy reward, applied
without reward scaling. GRPO, Dr.\ GRPO, and DAPO differ mainly in how they
treat the group standard deviation \citep{groupstd2026}. OCR rollouts are
near-deterministic, so within-group reward variance is often small and
dividing by it amplifies noise; a greedy-rollout baseline avoids the group
statistic.

The second concerns the reward itself. A binary pass-or-fail reward is
sparse, whereas graded partial credit supplies dense supervision from the
same deterministic checks \citep{wang2026verpo}, and every term we use is
graded. As
\autoref{tab:reward} lists, the total reward is a product of verifiable
terms: content similarity (normalized edit distance over mixed
\LaTeX{}/HTML), formula and table signals selected per dataset, structural
validity (brace balance, tag closure, table integrity), the fraction of
olmOCR-style unit tests that pass, and anti-repetition and format terms.
Structural, unit-test, and format terms are floored at 0.2 (the table term
at 0.1). Under a multiplicative composition a single failed check would
otherwise collapse the product to zero and remove the gradient signal from
an otherwise content-correct completion. The repetition term is not
floored, since degenerate loops are the failure mode that most readily
inflates the content score.

\begin{table}[t]
  \centering
  \caption{Multiplicative reward composition for GRPO. Enabled terms and
  floors vary by run.}
  \label{tab:reward}
  \footnotesize
  \resizebox{\textwidth}{!}{%
  \begin{tabular}{lll}
    \toprule
    Component & Signal & Role \\
    \midrule
    Content & Normalized edit distance on mixed \LaTeX{}/HTML & Textual fidelity \\
    Formula & Formula string matching & Formula correctness \\
    Table & TEDS, TEDS-S, table edit distance & Structure recovery \\
    Structural validity & Brace balance, tag closure, table integrity & Well-formedness \\
    Unit tests & Fraction of olmOCR-style presence, order, math, and table tests passed & Dense feedback \\
    Repetition and format & Repetition penalty, HTML conformance & Degeneration control \\
    \bottomrule
  \end{tabular}}
\end{table}

\subsection{Auxiliary regularizers}
\label{sec:regularizers}

Selected SFT and GRPO runs add auxiliary regularizers to the inherited MoE
decoder. Let $N$ be the number of routed experts, $T$ the number of
tokens, $p_{i,t}$ the router probability for expert $i$ on token $t$, and
$\mathbf{1}\{i{\in}\mathrm{top}k(t)\}$ its selection indicator. With
$f_i=\tfrac{1}{T}\sum_t\mathbf{1}\{i{\in}\mathrm{top}k(t)\}$ and
$P_i=\tfrac{1}{T}\sum_t p_{i,t}$, the sequence-level load-balancing loss is
$L_{\mathrm{bal}}=N\sum_i f_iP_i$ \citep{liu2024deepseekv3}. The normalized
gate-entropy term is
$L_{\mathrm{ent}}=\tfrac{1}{T\log N}\sum_t[-\sum_i p_{i,t}\log p_{i,t}]$.
We also use the output-logit penalty of PaLM \citep{chowdhery2022palm},
$L_z=\lambda_z\,\mathrm{logsumexp}(z)^2$.

Where token representations begin to collapse, we add a SimCTG-style
contrastive regularizer \citep{su2022simctg} with margin $m$:
\begin{equation}
  L_{\mathrm{ctr}}=\frac{1}{|\mathcal{P}|}
  \sum_{(i,j)\in\mathcal{P}}\max\bigl(0,m-(s_{ii}-s_{ij})\bigr),
\end{equation}
where $s_{ij}=\cos(h_i,h_j)$ and
$\mathcal{P}=\{(i,j):i\neq j,\ i,j\text{ are in the same local chunk}\}$.
The term encourages separation between nearby token states and follows the
regularization used in ReaderLM-v2 \citep{wang2025readerlmv2}.

\subsection{Agentic checkpoint merging and data curation}
\label{sec:merging}

\begin{figure}[t]
  \centering
  \begin{tikzpicture}[
    stage/.style={figbox, minimum width=3.7cm, minimum height=1.15cm},
    candidate/.style={figghost, minimum width=1.65cm, minimum height=0.42cm, font=\scriptsize},
    branch/.style={thick, blue!65},
  ]
    \node[stage] (train) at (0,0)
      {\textbf{Training Process}\\[-1pt] \scriptsize SFT and RLVR (GRPO / DAPO)};
    \node[stage] (merge) at (9.3,0)
      {\textbf{Agentic Model Merge}\\[-1pt] \scriptsize evaluate and synthesize outputs};
    \node[stage] (error) at (9.3,-3.0)
      {\textbf{Model Error Analysis}\\[-1pt] \scriptsize identify failures and edge cases};
    \node[stage] (curate) at (0,-3.0)
      {\textbf{Data Curation}\\[-1pt] \scriptsize filter and sample high-value data};

    \node[figtag, anchor=south west] at ($(train.north west)+(0,0.14)$) {Step 1};
    \node[figtag, anchor=south west] at ($(merge.north west)+(0,0.14)$) {Step 2};
    \node[figtag, anchor=south west] at ($(error.north west)+(0,0.14)$) {Step 3};
    \node[figtag, anchor=south west] at ($(curate.north west)+(0,0.14)$) {Step 4};

    \coordinate (split) at ($(train.east)+(0.4,0)$);
    \coordinate (join)  at ($(merge.west)+(-0.4,0)$);
    \node[candidate] (cand2) at (4.65,0)    {candidate 2};
    \node[candidate, above=0.2cm of cand2] (cand1) {candidate 1};
    \node[candidate, below=0.2cm of cand2] (candn) {candidate $N$};
    \node[figlbl, above=4pt of cand1] {generated candidates};

    \draw[branch] (train.east) -- (split);
    \draw[branch,dashed,rounded corners=3pt] (split) |- (cand1.west);
    \draw[branch] (split) -- (cand2.west);
    \draw[branch,dashed,rounded corners=3pt] (split) |- (candn.west);
    \draw[branch,dashed,rounded corners=3pt] (cand1.east) -| (join);
    \draw[branch] (cand2.east) -- (join);
    \draw[branch,dashed,rounded corners=3pt] (candn.east) -| (join);
    \draw[figflow] (join) -- (merge.west);
    \draw[figflow] (merge.south) -- node[figlbl, right=3pt] {merged candidate} (error.north);
    \draw[figflow] (error.west) -- node[figlbl, above=3pt] {diagnostic insights} (curate.east);
    \draw[figflow] (curate.north) -- node[figlbl, left=3pt] {curated dataset} (train.south);
  \end{tikzpicture}
  \caption{Outer training loop. SFT and GRPO runs produce candidate
  checkpoints, an agent merges them under an evaluation budget, and error
  analysis drives the next data mixture. FastMTP is fit on the selected
  verifier after the loop.}
  \label{fig:training}
\end{figure}
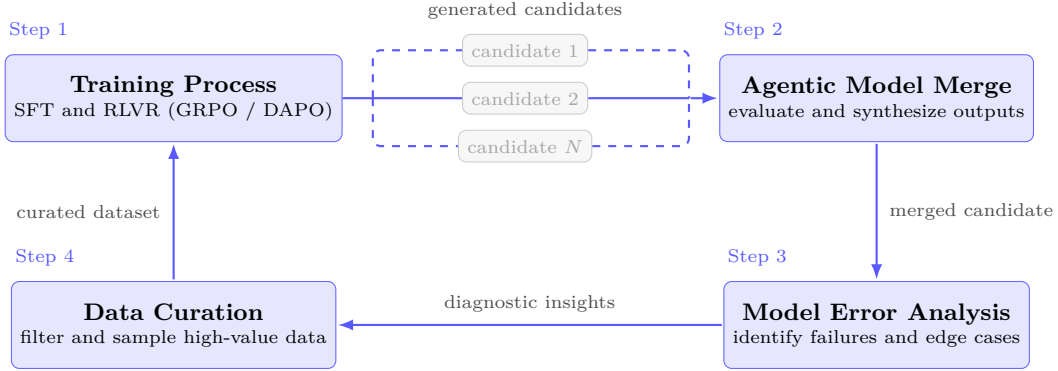

Alignment, robustness, and GRPO produce a pool of candidate checkpoints
that differ in data mixture, reward composition, and regularization. They
are applied repeatedly across the rounds of the outer loop in
\autoref{fig:training}: train candidates, merge, diagnose failures, and
curate data. The draft head is fitted once the loop selects a verifier, as
the next section describes.

Merging is a search problem. Under a fixed evaluation budget, an agent
evaluates merge configurations in parallel, scores them with unit-test and
edit-distance checks, and plans the next merges from previous outcomes.
The selected merge is then analyzed for errors, and high-value failure
modes are retrieved or synthesized by the methods of \autoref{sec:data}
and fed into the next round of alignment, robustness fine-tuning, or GRPO.

\subsection{Draft-head objective}
\label{sec:stage4}

After the outer loop selects a verifier, this stage fits a draft head to
that frozen model, using the state-shift alignment of
\autoref{sec:fastmtp}. For each valid response row $p$, define
$P_p^{(k)}=\operatorname{softmax}(\ell_p^{(k)})$ and the detached verifier
distribution
$Q_p=\operatorname{softmax}(W_{\mathrm{lm}}b_{p+1})$. Every recursive depth
uses the same row-aligned token and state targets:
\begin{equation}
\begin{aligned}
  L_{\mathrm{CE}}^{(k)}
    &= \mathbb{E}_{p}\!\left[-\log P_p^{(k)}(x_{p+2})\right],\\
  L_{\mathrm{KL}}^{(k)}
    &= \mathbb{E}_{p}\!\left[\mathrm{KL}\!\left(Q_p\,\|\,P_p^{(k)}\right)\right],\\
  L_{\mathrm{MSE}}^{(k)}
    &= \mathbb{E}_{p}\!\left[\left\|u_p^{(k)}-b_{p+1}\right\|_2^2\right].
\end{aligned}
\end{equation}
The expectations exclude prompt and padding positions. With
$\alpha_k=\beta^{k-1}/\sum_{r=1}^{K}\beta^{r-1}$ and $\beta=0.6$, the
objective is
\begin{equation}
  L = 0.85\sum_{k=1}^{K}\alpha_k L_{\mathrm{CE}}^{(k)}
      + 0.45\sum_{k=1}^{K}\alpha_k L_{\mathrm{KL}}^{(k)}
      + 0.45\sum_{k=1}^{K}\alpha_k L_{\mathrm{MSE}}^{(k)},
\end{equation}
where verifier states and distributions are detached. The trainable
parameters are $B_\theta$, $P$, $N_e$, and $N_h$. The verifier backbone
remains fixed, and the draft embedding, output norm, and LM head are tied
to the corresponding verifier weights and receive no updates.

\section{Evaluation}
\label{sec:evaluation}

\subsection{Benchmarks and metrics}
\label{sec:eval-bench}

We evaluate parsing quality and serving performance. olmOCR-Bench
\citep{poznanski2025olmocr} scores page-level parsing with unit-test-style
checks on text presence, reading order, math, and tables; we report the
official overall, including the headers/footers category. OmniDocBench
\citep{ouyang2025omnidocbench} reports text edit distance, formula CDM,
table TEDS/TEDS-S, and reading-order edit distance; we adopt the refined
v1.6 protocol, whose overall score averages the text score, formula CDM,
and table TEDS.

Baselines are recent general VLMs and specialized OCR models with a
published number under each protocol. Layout-plus-recognition pipelines are
omitted, and dots.mocr is retained as a two-pass specialized system. The
serving throughput reported in \autoref{tab:efficiency} uses a larger pool
on olmOCR-Bench under batched inference.

\subsection{Parsing quality}
\label{sec:eval-main}

\begin{table}[t]
  \centering
  \caption{olmOCR-Bench comparison. Baseline scores are as
  reported by LightOnOCR \citep{lightonocr}, the corresponding published
  works, and the {\href{https://huggingface.co/datasets/allenai/olmOCR-bench?leaderboard}{\texttt{leaderboard}}}.
  Params gives total and, for MoE models, active parameters. A dash means no
  published number under this protocol. Per-column best in bold.}
  \label{tab:olmocr}
  \footnotesize
  \setlength{\tabcolsep}{4pt}
    \resizebox{\textwidth}{!}{%
  \begin{tabular}{lcccccccccc}
      \toprule
      Model & Params & ArXiv & OldScans-Math & Tables & OldScans
      & Multi-col & LongTiny & Hdr/Ftr $^{\S}$ & Base & Overall $\uparrow$ \\
      \midrule
      \multicolumn{11}{l}{\textit{General VLMs$^{\ddagger}$}} \\
      Gemini 3 Flash & -- & 80.1 & 73.6 & 64.6 & 45.8 & 75.3 & 90.3 & 27.4 & -- & -- \\
      Qwen3-VL-235B & 235B/22B & 88.4 & 81.2 & 86.7 & 49.6 & \textbf{85.9} & 88.9 & 33.6 & -- & -- \\
      \midrule
      \multicolumn{11}{l}{\textit{Specialized OCR models}} \\
      DeepSeek-OCR & 3B/570M & 77.5 & 74.5 & 77.3 & 33.1 & 67.3 & 83.0 & \textbf{96.1} & 99.3 & 76.0$^{\|}$ \\
      dots.mocr & 3B & 85.9 & 85.5 & 90.7 & 48.2 & 85.3 & 81.6 & 94.0 & 99.7 & 83.9 \\
      olmOCR-2 & 8B & 82.9 & 82.1 & 84.3 & 48.3 & 84.3 & 81.4 & -- & 99.7 & 82.4 \\
      LightOnOCR-2 & 1B & \textbf{89.6} & 85.6 & 89.0 & 42.2 & 84.8 & 91.4 & 19.7$^{\dagger}$ & 99.6 & 83.2$^{\dagger}$ \\
      chandra-ocr-2 & 4B & 86.9 & \textbf{89.1} & \textbf{92.1} & \textbf{51.1} & 82.1 & \textbf{93.7} & 91.4 & \textbf{99.9} & \textbf{85.8} \\
      \midrule
      \jinaocr{} & 3B/570M & 86.1 & 82.3 & 88.8 & 42.6 & 85.5 & 93.2 & 88.7 & \textbf{99.9} & 83.4 \\
      \bottomrule
   \end{tabular}}
  \vspace{2pt}
  \begin{minipage}{\textwidth}\scriptsize
    $^{\dagger}$ LightOnOCR-2 excludes Hdr/Ftr from its overall; the Hdr/Ftr score is from Appendix~C.1 of its paper.
    $^{\ddagger}$ Per-subset scores from \texttt{idp-leaderboard.org}; the overall is omitted because it cannot be reproduced from the listed subsets.
    $^{\S}$ Text-absence tests that reward omitting headers and footers; models that transcribe all visible content score low, which explains the wide dispersion.
    $^{\|}$ Mean of the eight subsets as re-run by \citet{lightonocr}, whose published 73.1 excludes Hdr/Ftr.
  \end{minipage}
\end{table}

\autoref{tab:olmocr} reports olmOCR-Bench results.
\jinaocr{} reaches 83.4 overall, third among specialized models with a
published overall, behind chandra-ocr-2 (4B, 85.8) and the two-pass
dots.mocr (83.9), and ahead of LightOnOCR-2 (83.2$^{\dagger}$) and the 8B
olmOCR-2 (82.4). It is 7.4 points above DeepSeek-OCR (76.0), the backbone
it post-trains, which is the one comparison in the table that isolates our
post-training recipe. Among all listed systems it is second on multi-column (85.5) and
long-tiny text (93.2), and ties for best on Base (99.9).

\begin{table}[t]
  \centering
  \caption{OmniDocBench v1.6 comparison. Baseline scores are transcribed from
  the public leaderboard (OpenDataLab), except HunyuanOCR-1.5, which is
  taken from its technical report. Params gives total and, for MoE models,
  active parameters; RO denotes reading order. Per-column best in bold.}
  \label{tab:main}
  \footnotesize
  \setlength{\tabcolsep}{4.5pt}
  \resizebox{\textwidth}{!}{%
  \begin{tabular}{lccccccc}
    \toprule
    Method & Params & Overall $\uparrow$ & Text$_{\mathrm{Edit}}$ $\downarrow$
    & Formula$_{\mathrm{CDM}}$ $\uparrow$ & Table$_{\mathrm{TEDS}}$ $\uparrow$
    & Table$_{\mathrm{TEDS\mbox{-}S}}$ $\uparrow$ & RO$_{\mathrm{Edit}}$ $\downarrow$ \\
    \midrule
    \multicolumn{8}{l}{\textit{General VLMs}} \\
    Gemini 3 Flash & -- & 92.62 & 0.066 & 95.16 & 89.29 & 93.51 & 0.172 \\
    Qwen3-VL-235B & 235B/22B & 89.78 & 0.063 & 92.55 & 83.07 & 86.75 & 0.166 \\
    \midrule
    \multicolumn{8}{l}{\textit{Specialized OCR models}} \\
    DeepSeek-OCR-2 & 3B/570M & 90.25 & 0.050 & 91.84 & 83.89 & 87.75 & 0.144 \\
    HunyuanOCR-1.5 & 1B & 94.74 & 0.039 & 94.50 & 93.67 & 94.71 & 0.129 \\
    PaddleOCR-VL-1.6 & 0.9B & \textbf{96.34} & \textbf{0.033} & \textbf{97.53} & \textbf{94.76} & \textbf{97.10} & \textbf{0.128} \\
    \midrule
    \jinaocr{} & 3B/570M & 91.14 & 0.046 & 93.28 & 84.68 & 89.01 & 0.142 \\
    \bottomrule
  \end{tabular}}
\end{table}

On OmniDocBench, as \autoref{tab:main} reports, \jinaocr{} reaches 91.14
overall, third among specialized models with a published v1.6 score, behind
PaddleOCR-VL-1.6 (0.9B, 96.34) and HunyuanOCR-1.5 (1B, 94.74), and ahead of
DeepSeek-OCR-2 (90.25) and the much larger Qwen3-VL-235B (89.78). Its
margin over DeepSeek-OCR-2 holds on every column, including formula CDM
(93.28 against 91.84).

\subsection{Serving throughput}
\label{sec:eval-efficiency}

\begin{table}[t]
\centering
\caption{Serving efficiency of OCR models on olmOCR-Bench (1,403 pages, one
         A100 SXM4 40\,GB, concurrency 32). Score is the olmOCR-Bench overall from
         \autoref{tab:olmocr} or, for systems not listed there, from the public
         leaderboard; a dash means no published number. Bold is best, underline
         second best, per column.}
\label{tab:efficiency}
\begin{tabular}{lrrrr}
\toprule
Model & Score $\uparrow$ & Pages/s $\uparrow$ & Output tok/page $\downarrow$ & Output tok/s $\uparrow$ \\
\midrule
chandra-ocr-2          & \textbf{85.8} & 0.38 & 1917 &  730 \\
dots.mocr              & \underline{83.9} & 0.55 & 1711 &  934 \\
Surya OCR~2            & 83.3 & 1.05 & 3568 & \textbf{3760} \\
LightOnOCR-2           & 83.2$^{\dagger}$ & 1.33 & 1208 & 1606 \\
Infinity-Parser-7B     & 82.5 & 0.64 & 1066 &  680 \\
olmOCR-2               & 82.4 & 1.22 & 1128 & 1374 \\
MinerU2.5-Pro-1.2B     & 82.2 & 1.28 & 1548 & 1982 \\
DeepSeek-OCR           & 76.0 & \underline{2.10} & 1366 & \underline{2871} \\
GLM-OCR                & 75.2 & 1.21 & \underline{1054} & 1272 \\
MinerU2.5-1.2B         & 75.2 & 1.43 & 1563 & 2231 \\
PaddleOCR-VL-1.6       & -- & 0.33 & \textbf{1048} &  344 \\
HunyuanOCR-1.5         & -- & 0.83 & 1058 &  873 \\
DeepSeek-OCR-2         & -- & 1.50 & 1389 & 2083 \\
\midrule
\jinaocr{}             & 83.4 & \textbf{2.57} & 1085 & 2792 \\
\bottomrule
\end{tabular}
\end{table}

\begin{figure}[t]
  \centering
  \includegraphics[width=\textwidth]{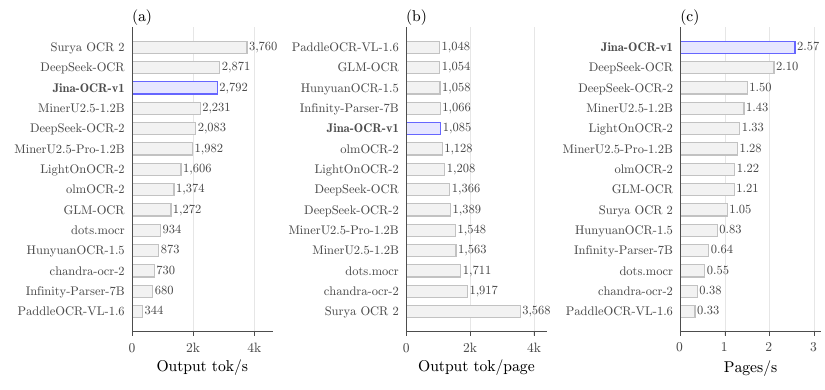}
  \caption{The pool of \autoref{tab:efficiency} ranked by (a)~output token
  throughput, (b)~output tokens per page, and (c)~page throughput; blue
  marks this work.}
  \label{fig:barplots}
\end{figure}

We measure the larger pool of \autoref{tab:efficiency} under the same
olmOCR-Bench serving setup as panel~(b) of \autoref{fig:overview}, and
\autoref{fig:barplots} ranks it by each quantity. Token
throughput alone does not determine page throughput: Surya
OCR~2 leads in tok/s ($3{,}760$) but emits $3{,}568$ tokens per page,
whereas \jinaocr{} emits $1{,}085$ and finishes pages faster.

The two are related by
$\text{pages/s}=(\text{tok/s})/(\text{tok/page})$, and across
\autoref{tab:efficiency} they are positively correlated ($r{=}{+}0.55$
between $\log$ tok/s and $\log$ tok/page, $n{=}13$): faster-decoding
systems also emit longer transcriptions, so the two effects partly cancel
and the rankings diverge. Output length is independent of parsing quality
($r{=}{+}0.22$ across the $n{=}10$ systems with a published overall).
Conciseness can therefore be optimized on its own, and \jinaocr{} reaches
the highest page throughput by combining competitive token throughput with
the shortest outputs of any system scoring above 83.

\subsection{Speculative decoding}
\label{sec:eval-mtp}

\autoref{tab:efficiency_olmocr} reports speculative-decoding efficiency on
olmOCR-Bench with vLLM~0.20.1 at batch size 1 on a low-budget GPU, the
NVIDIA L4. We compare
greedy autoregressive decoding ($k{=}0$) with FastMTP at $k\in\{1,2,3\}$
in eager and CUDA-graph execution. FastMTP at $k{=}3$ reaches $1.95\times$
the eager autoregressive baseline. CUDA graphs raise that baseline from
$42.7$ to $158.3$ tok/s, a factor of $3.71$, and against it the best depth
is $k{=}1$ at $1.17\times$.

Writing the speedup as $S=\tau/c$, with $\tau$ and $c$ as defined in
\autoref{tab:efficiency_olmocr}, separates draft quality from step cost.
Acceptance is mode-invariant: $\tau$ is $2.73$ in eager and $2.74$ in graph
mode at $k{=}3$. The cost $c$ grows from $1.40$ to $2.51$, which is the same
absolute overhead of about 9~ms per speculative step in both modes, so the
speedup scales with the cost of the verifier step it replaces. Conditional
acceptance falls from $0.83$ at the first draft position to $0.65$ at the
third, so each additional depth contributes less while adding a fixed cost;
the best depth is $k{=}3$ in eager mode and $k{=}1$ under CUDA graphs.

\begin{table}[t]
  \centering
  \caption{Decoding efficiency of \jinaocr{} with FastMTP on olmOCR-Bench
  on a low-budget GPU (NVIDIA L4, vLLM 0.20.1, batch size 1). Speedup $S$ is
  against the $k{=}0$ baseline of the same execution mode. Acceptance rate is
  the position-averaged $(\tau-1)/k$, where $\tau$ is the mean number of tokens
  committed per speculative step including the bonus token, $c=\tau/S$ is the
  cost of one speculative step in units of one autoregressive step, and $a_d$
  is the probability that draft depth $d$ is accepted given that all shallower
  depths were. Bold marks the best operating point within each execution mode.
  Values are measured at batch size 1 on a different device from
  \autoref{tab:efficiency} and are not comparable with it.}
  \label{tab:efficiency_olmocr}
  \small
  \begin{tabular}{llccccc}
    \toprule
    Mode & $k$ & Output tok/s $\uparrow$ & Speedup $S$ $\uparrow$ & Acceptance rate & $\tau$ & $c$ $\downarrow$ \\
    \midrule
    \multirow{4}{*}{Eager}
      & 0 & 42.7  & 1.00$\times$ & --     & --   & 1.00 \\
      & 1 & 64.0  & 1.50$\times$ & 82.6\% & 1.83 & 1.22 \\
      & 2 & 77.9  & 1.82$\times$ & 69.1\% & 2.38 & 1.30 \\
      & 3 & \textbf{83.1} & \textbf{1.95}$\times$ & 57.6\% & 2.73 & 1.40 \\
    \midrule
    \multirow{4}{*}{Graph}
      & 0 & 158.3 & 1.00$\times$ & --     & --   & 1.00 \\
      & 1 & \textbf{185.6} & \textbf{1.17}$\times$ & 82.9\% & 1.83 & 1.56 \\
      & 2 & 183.8 & 1.16$\times$ & 69.3\% & 2.38 & 2.05 \\
      & 3 & 172.9 & 1.09$\times$ & 57.9\% & 2.74 & 2.51 \\
    \bottomrule
  \end{tabular}

  \vspace{2pt}
  {\footnotesize
  \begin{tabular}{lccc}
    \toprule
    Conditional acceptance $a_d$ & $d{=}1$ & $d{=}2$ & $d{=}3$ \\
    \midrule
    Eager & 0.830 & 0.663 & 0.636 \\
    Graph & 0.830 & 0.663 & 0.655 \\
    \bottomrule
  \end{tabular}}
\end{table}

\section{Conclusion}
\label{sec:conclusion}

We presented \jinaocr{}, an end-to-end document parser built on the
compressed-vision, 3B-MoE backbone of DeepSeek-OCR, with a recursively
shared FastMTP draft head and post-training under dense verifiable rewards.
It scores $83.4$ on olmOCR-Bench and $91.14$ on OmniDocBench v1.6, and it
is the fastest system in pages per second among those we measure. Page
throughput and token throughput rank systems differently, because
faster-decoding systems also emit longer transcriptions, and output length
is independent of parsing quality, so conciseness can be optimized on its
own. On a low-budget GPU such as the NVIDIA L4, the FastMTP head nearly
doubles decoding speed, with acceptance behavior that is invariant to the
execution mode of the verifier.

\bibliographystyle{plainnat}
\bibliography{references}

\end{document}